\documentclass[letterpaper, 10 pt, conference]{ieeeconf}  

\IEEEoverridecommandlockouts                              

\usepackage{graphicx} 
\usepackage{epsfig} 
\usepackage{mathptmx} 
\usepackage{times} 
\usepackage{amsmath} 
\usepackage{amssymb}  
\usepackage{amsfonts}
\usepackage[noadjust]{cite}
\usepackage{textcomp}
\usepackage[usenames, dvipsnames]{xcolor}
\usepackage[noend]{algpseudocode}
\usepackage{algorithm,algorithmicx}
\usepackage{hyperref}
\hypersetup{
  colorlinks=true,
  linkcolor=blue,
  urlcolor=blue,
}

\newlength\myindent
\title{\LARGE \bf
Stochastic Planning for ASV Navigation Using Satellite Images}

\author{Yizhou Huang$^{1}$, Hamza Dugmag$^{2}$, Timothy D. Barfoot$^{2}$, and Florian Shkurti$^{1}$ 
\thanks{$^{1}$Department of Computer Science, University of Toronto, 40 St George St, Toronto, Canada
        {\tt\small \{phuang, florian\}@cs.toronto.edu}}%
\thanks{$^{2}$University of Toronto Institute for Aerospace Studies, 4925 Dufferin St, Toronto, Canada
        {\tt\small hamza.dugmag@mail.utoronto.ca, tim.barfoot@utoronto.ca}}%
}

\begin{document}

\maketitle
\thispagestyle{empty}
\pagestyle{empty}

\begin{abstract}
Autonomous surface vessels (ASV) represent a promising technology to automate water-quality monitoring of lakes. 
In this work, we use satellite images as a coarse map and plan sampling routes for the robot. 
However, inconsistency between the satellite images and the actual lake, 
as well as environmental disturbances such as wind, aquatic vegetation, and changing water levels can make it difficult for robots to visit places suggested by the prior map. 
This paper presents a robust route-planning algorithm that minimizes the expected total travel distance given these environmental disturbances, which induce uncertainties in the map. 
We verify the efficacy of our algorithm in simulations of over a thousand Canadian lakes and demonstrate an application of our algorithm in a 3.7 km-long real-world robot experiment on a lake in Northern Ontario, Canada. Videos are available on our \href{https://pcctp.github.io/}{website}.
\end{abstract}

\section{Introduction}


Autonomous Surface Vessels (ASVs) have seen increasing attention as a technology to monitor rivers, lakes, coasts, and oceans in recent years \cite{robot_env2012, Odetti2020-iv, Ferri2015-sb,Madeo2020-yq,Cao2020-fc,Dash2021-uf,Ang2022-ep,MahmoudZadeh2022-su}. A fundamental challenge to the wide adoption of ASVs is the ability to navigate safely and autonomously in uncertain environments, especially for long durations. 
For example, a robot may precompute a plan that visits certain target areas on a map and execute it online. 
However, disturbances such as strong winds, waves, aquatic plants, unseen obstacles, and even simply changing visual appearances in a water environment are challenging for ASV navigation. 
Many potential low-level failures in robot perception and control systems may also undermine the overall success of the mission. 

Our goal is to use an ASV to monitor lake environments and collect water samples for scientists. 
To make the system robust, we propose to identify uncertainties (e.g., obstacles, wind-prone areas) that could block waterways, incorporate them into the map, and generate stochastic policies that can adapt online. 
One planning framework that is suitable for modelling uncertain paths is the Canadian Traveller Problem (CTP) \cite{ctp1991}. 
The most significant feature in a CTP graph is the stochastic edge, which has a probability of being blocked.
The state of any stochastic edge can be disambiguated by visiting the edge. 
Once the state has been classified as traversable or not, it remains the same.

In this paper, we propose a navigation framework --- the Partial Covering Canadian Traveller Problem (PCCTP) --- to solve a route-planning problem in an uncertain environment.
The framework uses a stochastic graph derived from coarse satellite images to plan an adaptive policy that visits all reachable target locations. 
Stochasticity in the graph represents possible events where a water passage between two points is blocked due to changing water levels, strong wind, and other unmapped obstacles.

Our proposed solution computes the optimal policy offline with a best-first tree-search algorithm. 
The benefits of our proposed approach are twofold:
1) Computing an optimal policy can reduce the total distance to visit all targets and return, 
2) The execution of our stochastic policy is robust even in the presence of environmental disturbances and partial robot system failure. 
We evaluate our solution method on a random set of Canadian lakes and demonstrate its robustness on a robot system that we built for field tests.

\begin{figure}[t!]
    \centering
    \includegraphics[width=\columnwidth]{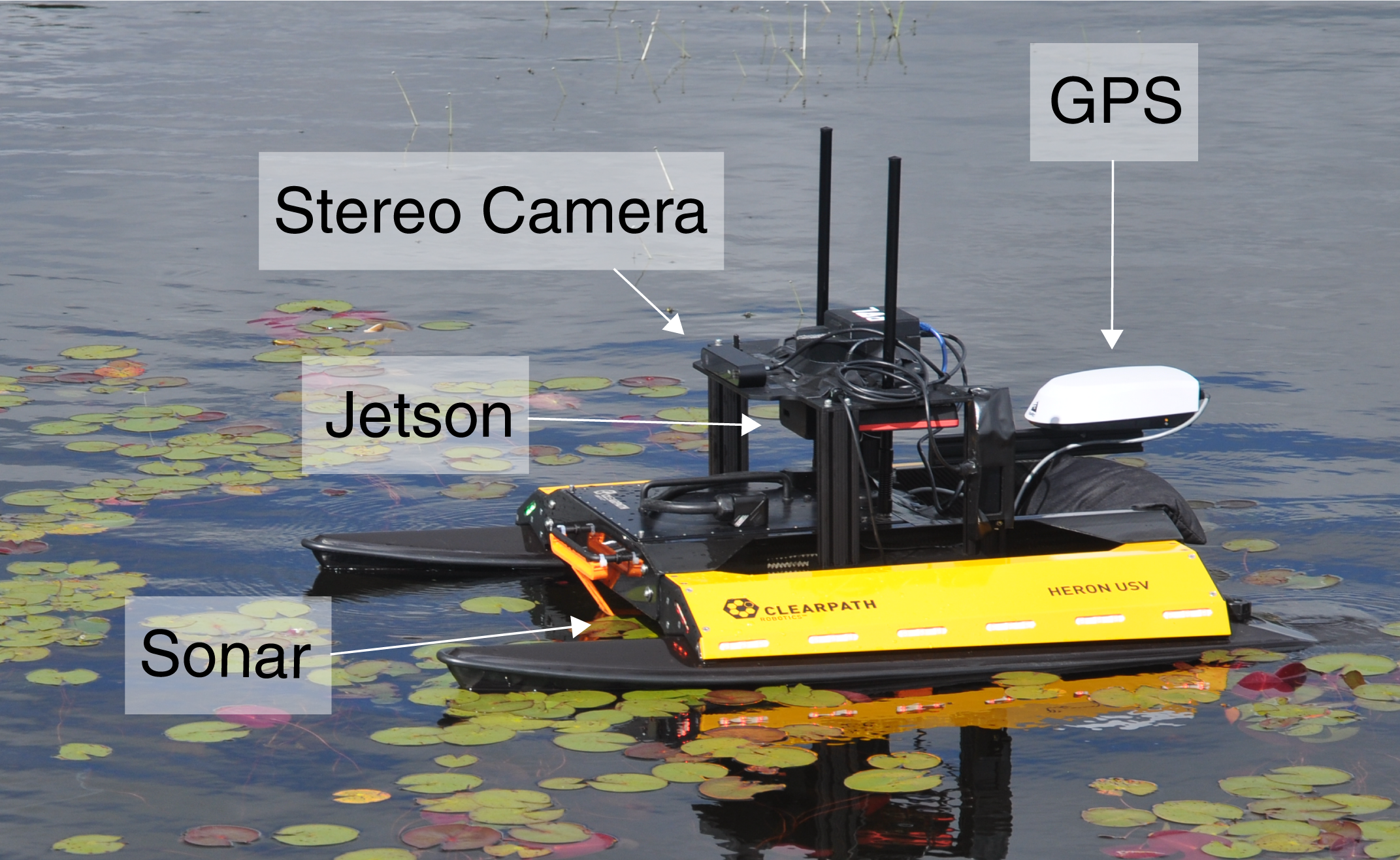}
    \caption{Our \textit{Clearpath Heron} ASV for water-quality monitoring during a field test. The ASV has onboard sensors (GPS, IMU, underwater scanning sonar, stereo camera) and an \textit{Nvidia Jetson} to process sensor measurements.}
    \label{fig:robot}
    \vspace{-0.6cm}
\end{figure}

\vspace{-0.1cm}
\section{Related Works}
\vspace{-0.1cm}
Remote sensing is a popular technique to build maps and monitor changes in water bodies around the world because of its efficiency \cite{WaterRemoteSensing, Yang2017-vb}.
The \textit{JRC} Global Surface Water dataset \cite{JRCwater} maps changes in water coverage from 1984 to 2015 at a 30 m by 30 m resolution, produced using \textit{Landsat} satellite imagery. 
Since water has a lower reflectance in the infrared channel, an effective method is to calculate water indices, such as NDWI \cite{ndwi} or MNDWI \cite{mndwi}, from two or more optical bands (e.g., green and near-infrared).
However, extracting water data using a threshold in water indices can be non-trivial due to variations introduced by clouds, seasonal changes, and sensor-related issues. 
To address this, \cite{Li2012-bn} and \cite{Feyisa2014-io} have developed techniques to select water-extraction thresholds adaptively.
Our approach aggregates water indices from historical satellite images to estimate probabilities of water coverage (see Sec. \ref{sec:graph_estimation}).
Overall, we argue that it is beneficial to build stochastic models of surface water bodies due to their dynamic nature and imperfect knowledge derived from satellite images. 

\begin{figure}[t!]
    \centering
    \includegraphics[width=0.9\columnwidth]{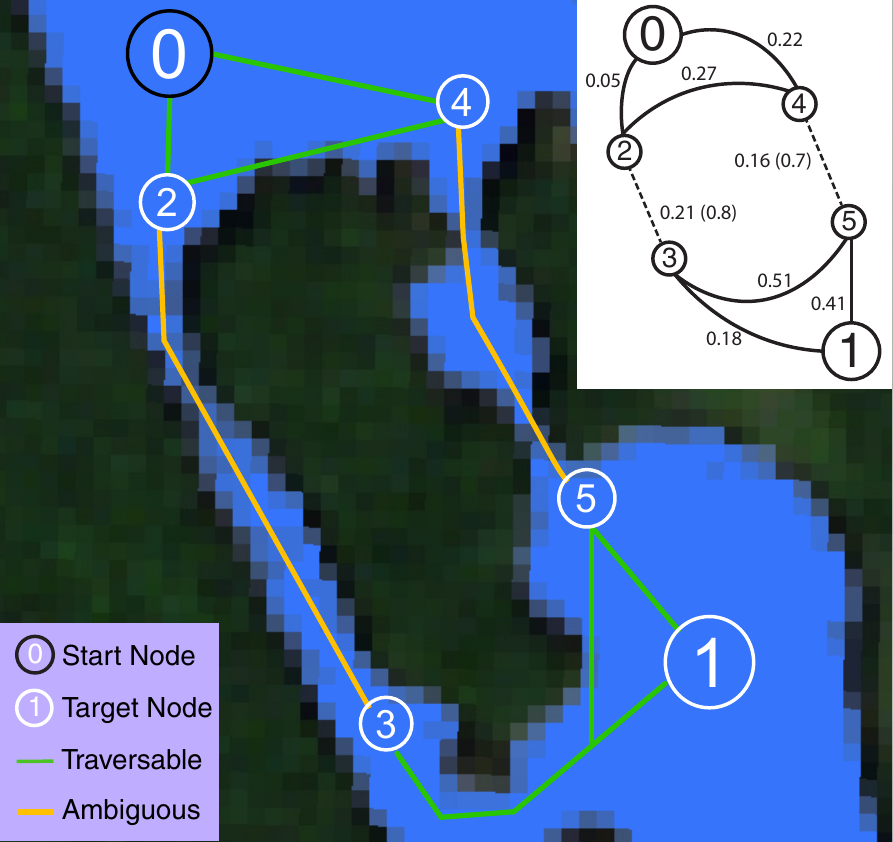}
    \caption{A toy example graph shown on a satellite image.
    The planned paths between nodes are simplified for ease of understanding. 
    The number beside each edge of the top-right graph is the path length in km, and the number in brackets is the blocking probability, which is computed using the probability of water coverage in each pixel (represented by its shade of \textcolor{CornflowerBlue}{blue}) on the path.
    Note that traversable and ambiguous edges are the state before any action.}
    \vspace{-0.6cm}
    \label{fig:example_graph}
\end{figure}

The other significant pillar of building an ASV navigation system is route planning. First formulated in the 1930s, the Travelling Salesperson Problem (TSP) \cite{TSP_survey} studies how to find the shortest path in a graph that visits every node once and returns to the starting node. 
Modern TSP solvers such as the \textit{Google} OR-tools \cite{ortools} can produce high-quality approximate solutions for graphs with about 20 nodes in a fraction of a second. 
In many cases, the problem graphs are built from real-world road networks and the edges are assumed to be always traversable. 
In CTP \cite{ctp1991}, however, edges can be blocked with some probability.
The goal is to compute a policy that has the shortest expected path to travel from a start node to a single goal node. 
CTP can also be formulated as a Markov Decision Process \cite{bellman1957markovian} and solved optimally with dynamic programming \cite{Polychronopoulos_undated-rw} or heuristic search \cite{ctpao*}.
The robotics community has also studied ways in which the CTP framework can be best used in path planning \cite{Ferguson2004-cf, ctp-guo2019}. 
Our problem setting, PCCTP, lies at the intersection of TSP and CTP, where the goal is to visit a partial set of nodes on a graph with stochastic edges. 
A similar formulation, known as the Covering Canadian Traveller Problem (CCTP) \cite{cctp}, presents a heuristic, online algorithm named Cyclic Routing (CR) to visit every node in a complete $n$-node graph with at most $n-2$ stochastic edges.
A key distinction between CCTP and our setting is that CCTP assumes all nodes are reachable, whereas the robot may give up on unreachable nodes located behind an untraversable edge in PCCTP.

In recent years, more ASV systems and algorithms for making autonomous decisions to monitor environments have been built. 
Schiaretti et al. \cite{Schiaretti2017-tl} classify the autonomy level for ASVs into 10 levels based on control systems, decision-making, and exception handling. 
Many works consider the mechanical, electrical, and control subsystems of their ASV designs \cite{Ang2022-ep,Madeo2020-yq,Ferri2015-sb}. 
Dash et al. \cite{Dash2021-uf} validated the use and accuracy of deploying ASVs for water-quality modelling by comparing the data collected from ASVs against independent sensors.
Two examples of vertically integrated autonomous water-quality monitoring systems using ASVs are presented in \cite{Chang2021-jz} and \cite{Cao2020-fc}.
In contrast, our main contribution is a robust mission-planning framework that is complementary to existing designs of ASV systems.
Finally, informative path planning is another orthogonal area where the robot relies on a probabilistic model to identify targets that maximize information gain; \cite{Bai2021-su} reviews this topic.

\section{Methodology}

\subsection{The Problem Formulation}
\label{subsection: problem}
We are interested in planning on a graph representation of a lake where parts of the water are stochastic (i.e., traversability is uncertain). 
Constructing such a graph using all pixels of satellite images is impractical since images are very high-dimensional. 
Thus, we extend previous works from CTP \cite{ctp1991, cctp, ctp-guo2019} 
and distill satellite images into a high-level graph $G$ where some stochastic edges $e$ may be untraversable with probability $p$. 
The state of a stochastic edge can be disambiguated only when the robot traverses the edge in question. 
The robot begins at the starting node $s$ and is tasked to visit all reachable targets $J$ specified by the user (e.g., scientists) before returning to the starting node. 
If some target nodes are unreachable because some stochastic edges block them from the starting node, the robot may give up on these sampling targets.
Hence we call this problem PCCTP.
The state of the robot is defined as a collection of the following: a list of target nodes that it has visited, the current node it is at, and its knowledge about the stochastic edges. 
A policy sets the next node to visit given the current state of the robot. 
The objective is to find the optimal policy $\pi^*$ that minimizes the expected cost to cover all reachable targets. 
Formally, we define the following terms:
\begin{itemize}
    \item $G = (V, E)$ is an undirected graph.
    \item $c : E \rightarrow \mathbb{R}_{\geq 0}$ is the cost function for an edge, which is the length of the shortest waterway between two points.
    \item $p : E \rightarrow [0, 1]$ is the blocking probability function. 
    \item $k$ is the number of stochastic edges.
    \item $s \in V$ is the start and return node.
    \item $J \subseteq V$ is the subset of target nodes to visit.
    \item $I = \{ \text{A}, \text{T}, \text{U} \}^k$ is an information vector that represents the robot's knowledge of the status of all $k$ stochastic edges.
    A, T, and U stand for ambiguous, traversable, and untraversable, respectively.
    \item $S \subseteq J$ is the subset of target nodes that have been visited.
    \item $a$ is the current node the robot is at.
    \item $x = (a, S, I)$ is the state of the robot.
    \item $\pi^*$ is the optimal policy that minimizes the cost $\mathbb{E} \left[\phi \left(\pi \right)\right]$, where $\phi$ is cost functional of the policy $\pi$. 
\end{itemize}

\begin{figure}[t!]
    \centering
    \includegraphics[width=\columnwidth]{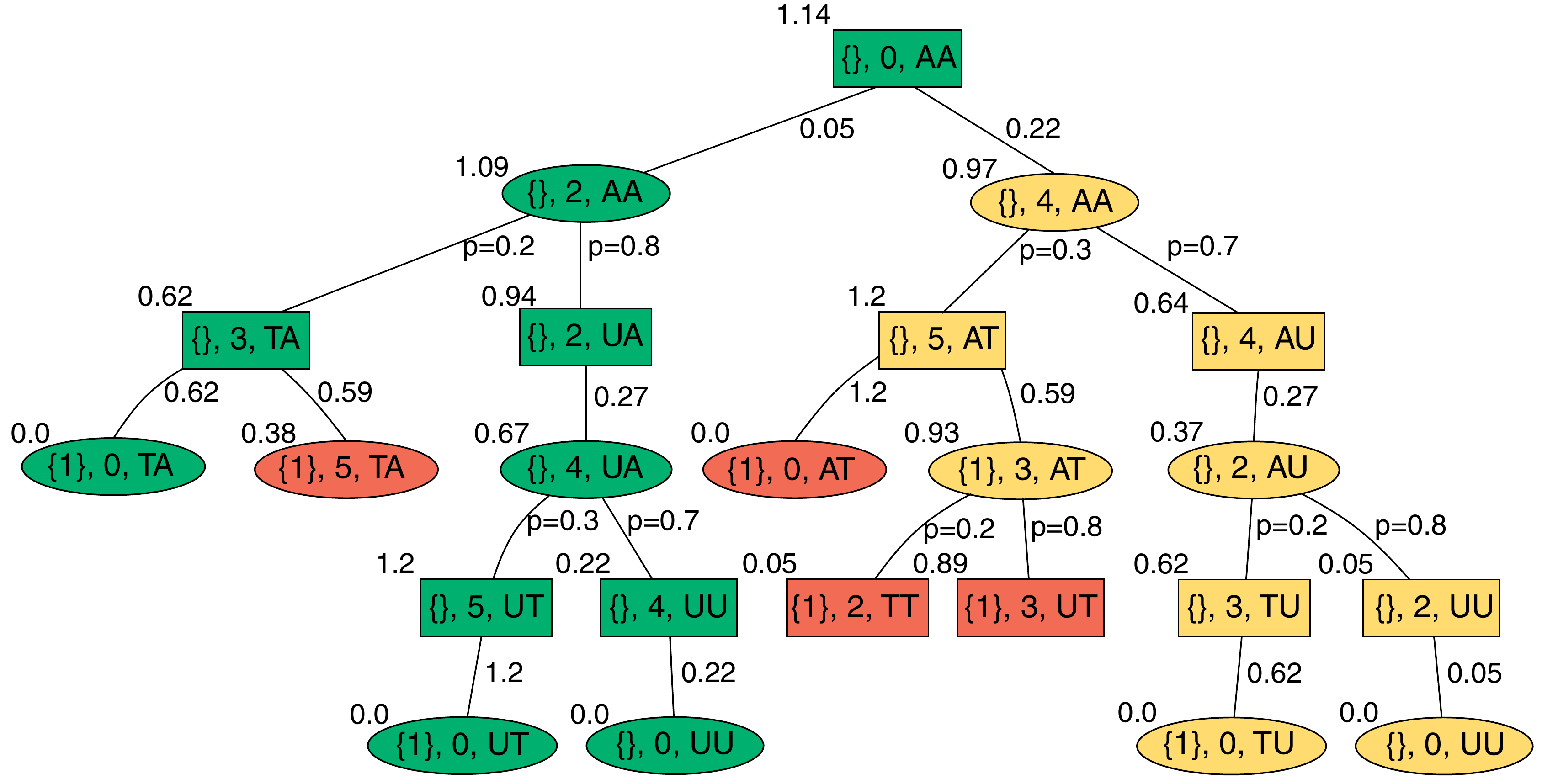}
    \caption{The final AO tree after running PCCTP-AO* on the example in Fig. \ref{fig:example_graph}. 
    The label inside each node is the current state of the robot.
    OR nodes are squares and AND nodes are ellipses. 
    Nodes that are part of the final policy are \textcolor{PineGreen}{green}, 
    extra expanded nodes are \textcolor{Goldenrod}{yellow}, 
    and leaf nodes terminated early are \textcolor{BrickRed}{red}.
    Some \textcolor{BrickRed}{red} nodes that are terminated early are left out in this figure for simplicity.}
    \label{fig:ao_tree}
\vspace{-0.4cm}
\end{figure}

\subsection{Solving PCCTP with AO*}

We extend the AO* search algorithm \cite{ctpao*} used in CTP to find exact solutions to our problem. 
AO* is a heuristic, best-first search algorithm that iteratively builds an AO tree to explore the state space until the optimal solution is found. 
In this section, we will first explain how to use an AO tree to represent a PCCTP instance, then break down how to use AO* to construct the AO tree containing the optimal policy.

\textbf{AO Tree Representation of PCCTP} The construction of the AO tree is a mapping of all possible actions the robot can take and all possible disambiguation outcomes at every stochastic edge. 
Following \cite{ctpao*}, an AO tree is a rooted tree $T = (N, A)$ with two types of nodes (OR node or AND node) and arcs. 
The node set $N$ can be partitioned into the set of OR nodes $N_O$ and the set of AND nodes $N_A$. 
An OR node lists all possible actions the robot can take next, whereas an AND node corresponds to a disambiguation event in which the robot needs to consider all possible outcomes.
Each arc in $A$ represents either an action or a disambiguation outcome, and is not the same as $G$'s edges ($A \neq E$). 

Specifically, each node $n$ is assigned a label $(n.a, n.S, n.I)$ that represents the state of the robot. 
$n.a$ is the current node, $n.S$ is the set of visited targets, and $n.I$ is the information vector containing the current knowledge of the stochastic edges.
The root node $r$ is an OR node with the label $(s, \emptyset, \mathrm{AA...A})$, representing the starting state of the robot.
An outgoing arc from an OR node $n$ to its successor $n'$ represents an action, which can be either visiting the remaining targets and returning to the start, or going to the endpoint of an ambiguous edge via some target nodes along the way. 
An AND node has two successors describing both possible outcomes of the disambiguation event of a stochastic edge.
Each succeeding node of an OR node is either an AND node or a leaf node.
A leaf node means the robot has visited all reachable target nodes and has returned to the start node.

Each arc $(n, n')$ is assigned a cost $c$, which is the length of travelling from node $n.a$ to node $n'.a$ while visiting the subset of newly visited targets $n'.S \setminus n.S$ along the way.
For all outgoing arcs of an AND node, a function $p : A \to [0, 1]$ assigns the traversability probability for the stochastic edge.
The cost of disambiguating that edge is simply its length.
The cost-to-go function $f : N \to \mathbb{R}_{\geq 0}$ is the function that satisfies the following conditions:
\begin{itemize}
    \item if $n \in N_A$, $f(n) =  \sum_{n' \in N(n)} [p(n, n') \times (f(n') + c(n, n'))]$,
    \item if $n \in N_O$, $f(n) = \min_{n' \in N(n)} [f(n') + c(n, n')]$,
    \item if $n \in N$ is a leaf node, $f(n) = 0$.
\end{itemize}

\setlength{\textfloatsep}{2pt}
\begin{algorithm}[t!]
    \caption{The PCCTP-AO* Algorithm}
    \label{algo:pcctp}
    \begin{algorithmic}[1]
    \Require {$G$, $c$, $p$, $J$, $k$, $s$}
    \State $n.a = s$, $n.S = \emptyset$, $n.I = \{A\}^k$
    \State $f(n) = h(n)$; $n$.type = OR; $T.$root = $n$
    \While {$T.$root.status $\neq$ solved}
        \State {$n \gets$ \Call{SelectNode}{$T.$root}}
        \For {$n' \in \Call{Expand}{n, T}$}
            \State $f(n') = h(n')$
            \If {$\Call{ReachableSet}{J, \; n'.I} \subseteq n'.S$}
            \State $n'.$status = solved
            \EndIf
        \EndFor
        \State \Call{Backprop}{$n, T$}
    \EndWhile
    \State {\Return $T$}
    
    \Function {SelectNode}{$n$} \Comment{Find the most promising subtree recursively until reaching a leaf node.}
    
    \EndFunction
    \Function {Expand}{$n$, $T$} \Comment{Find the set of succeeding nodes for node $n$ and it add to the tree $T$.}
    \EndFunction
    \Function {Backprop}{$n, T$} \Comment{Update the cost of the parent of $n$ recursively until the root. Same as in \cite{ctp-guo2019}.
    }
    \EndFunction

    \end{algorithmic}
\end{algorithm}

Once the complete AO tree is constructed, the optimal policy is the collection of nodes and arcs that are included in the calculation of the cost-to-go from the root of the tree, 
and the optimal expected cost is simply $f(r)$.
For example, the optimal action at an OR node $n$ is the arc $(n, n')$ that minimizes the cost-to-go from $n$, while the next action at an AND node depends on the disambiguation outcome.
However, constructing the full AO tree from scratch is not practical since the space complexity is exponential with respect to the number of stochastic edges. 
Instead, we use the heuristic-based AO* algorithm, explained below. 

\textbf{PCCTP-AO* Algorithm}
Our PCCTP-AO* algorithm (Algorithm \ref{algo:pcctp}) is largely based on the AO* algorithm \cite{ao*1971, ao*1978}. 
AO* utilizes an admissible heuristic $h : N \to \mathbb{R}_{\geq 0}$ that underestimates the cost-to-go $f$ to build the AO tree incrementally from the root node until the optimal policy is found.
The algorithm expands the most promising node in the current AO tree based on a heuristic, and backpropagates its parent's cost recursively to the root.
This expansion-backpropagation process is repeated until the AO tree includes the optimal policy.

One key difference between AO* and PCCTP-AO* is that the reachability of a target node may depend on the traversability of a set of critical stochastic edges connecting the target to the root.
For example, the two stochastic edges in the top-right graph of Fig. \ref{fig:example_graph} are critical because target node 1 would be unreachable if both edges are blocked. 
Thus, a simple heuristic that assumes all ambiguous edges are traversable may overestimate the cost-to-go if skipping unreachable targets reduces the overall cost.

Alternatively, we can construct the following relaxed problem to calculate the heuristic. 
If a stochastic edge is not critical to any target, we still assume it is traversable.
Otherwise, we remove the potentially unreachable target for the robot and instead, disambiguate one of the critical edges of the removed target.
The heuristic is the cost of the best plan that covers all definitively reachable targets and disambiguates one of the critical stochastic edges. 
For example, consider computing the heuristic at starting node 0 in Fig. \ref{fig:example_graph}. 
The optimistic plan is to visit node 2, disambiguate the edge (2, 3), and return to node 0.
If the robot reaches node 3, the heuristic would then be the remaining cost to visit node 1 and return to the start.
This heuristic is always admissible because the path to disambiguate a critical edge is always a subset of the eventual policy.
We can compute this by constructing an equivalent generalized travelling salesman problem \cite{set-tsp} and solve it with any optimal TSP solver.

Fig. \ref{fig:ao_tree} shows the result of applying PCCTP-AO* to the example problem in Fig. \ref{fig:example_graph}.
Note that the AO* algorithm stops expanding as soon as the lower bound of the cost of the right branch exceeds that of the left branch.
This guarantees the left branch has the lower cost and thus, is optimal.

\subsection{Estimating Stochastic Graphs From Satellite Imagery}
\label{sec:graph_estimation}
We will now explain our procedure to estimate the high-level stochastic graph from satellite images.

\textbf{Water Masking} Our goal in the first step is to build a water mask of a water area across a specific period of time (e.g., 30 days).
We use the \textit{Sentinel-2} Level 2A dataset \cite{sentinel-2}, which has provided images at 10 m by 10 m resolution since 2017.
Each geographical location is revisited every five days by a satellite. 
We then select all satellite images in the target spatiotemporal window and filter out the cloudy images using the provided cloud masks.
For each image, we calculate the Normalized Difference Water Index (NDWI) \cite{ndwi} for every pixel using green and near-infrared bands.
However, the distribution of NDWI values varies significantly across different images over time.
Thus, we separate water from land in each image and aggregate the indices over time.
We then fit a bimodal Gaussian Mixture Model on the histogram of NDWIs to separate water pixels from non-water ones for each image.   
We average all water masks over time to calculate the probabilistic water mask at the target spatiotemporal window.
Each pixel on the final mask represents the probability of water coverage on this 10 m by 10 m area.
If a pixel may or may not have water, then we call it a stochastic pixel.
Finally, we identify the boundary of all deterministic water pixels.

\textbf{Stochastic Edge Detection: Pinch Point} We can now identify those stochastic water paths (i.e., narrow straits, pinch points \cite{Ferguson2004-cf}) that are useful for navigation. 
A pinch point (e.g., Fig. \ref{fig:example_graph}) is defined as a sequence of stochastic water pixels that connect two parts of topologically far (or distinct) but metrically close water areas.
Essentially, this edge is a shortcut connecting two points on the water boundary that are otherwise far away or disconnected. 
To find all such edges, we iterate over all boundary pixels, test each shortest stochastic water path to nearby boundary pixels, and include those stochastic paths that are shortcuts.
The blocking probability of a stochastic edge is one minus the minimum water probability along the path. 
Since this process will produce many similar stochastic edges around the same narrow passage, we run DBSCAN \cite{dbscan} and only choose the shortest stochastic edge within each cluster. 

\textbf{Stochastic Edge Detection: Windy Edges} The second type of stochastic edges are those with strong wind. In practice, when an ASV travels on a path far away from the shore, there is a higher chance of running into a strong headwind or wave, making the path difficult to traverse. We define an edge to be a windy edge if it is 200 m away from the water boundary at some point, and assign a small probability for the event where the edge is blocked by wind. 

\textbf{Path Generation} The next step is to construct the geo-tagged path and calculate all edge costs in the high-level graph.
The nodes in the high-level graph are composed of all sampling targets, endpoints of stochastic edges, and the starting node. 
We run A* \cite{hart1968formal} on the deterministic water pixels to calculate the shortest path between every pair of nodes except for the stochastic edges found in the previous step.
Since the path generated by A* connects neighbouring pixels, we smooth them by downsampling.
Then, we can discard any unnecessary stochastic edges if they do not reduce the distance between a pair of nodes.
Finally, we check if each deterministic edge is a windy edge, and obtain the high-level graph used in PCCTP.

\section{Simulations}
\vspace{-0.1cm}

\subsection{Testing Dataset}
\vspace{-0.1cm}
We evaluate our route-planning framework on Canadian lakes selected from the \textit{CanVec Series} Ontario dataset \cite{canvec}. Published by \textit{Natural Resources Canada}, this dataset contains geospatial data of over 1.1 million water bodies in Ontario.
Considering a practical mission length, lakes are filtered such that their bounding boxes are 1-10 km by 1-10 km.
Then, water masks of the resulting 5190 lakes are generated using \textit{Sentinel-2} imagery across 30 days in June 2018-2022 \cite{sentinel-2}.
We then detect any pinch points on the water masks and randomly sample five different sets of target nodes on each lake, each with a different number of targets. The starting locations are sampled near the shore. 

Furthermore, we generate the high-level graphs as well as windy edges from the water mask. 
Graphs with no stochastic edges are removed as well as any instances with more than nine stochastic edges due to long run times.
In the end, we evaluate our algorithm on 2217 graph instances, which come from 1052 unique lakes. 

\begin{figure}[t!]
    \centering
    \includegraphics[width=\columnwidth]{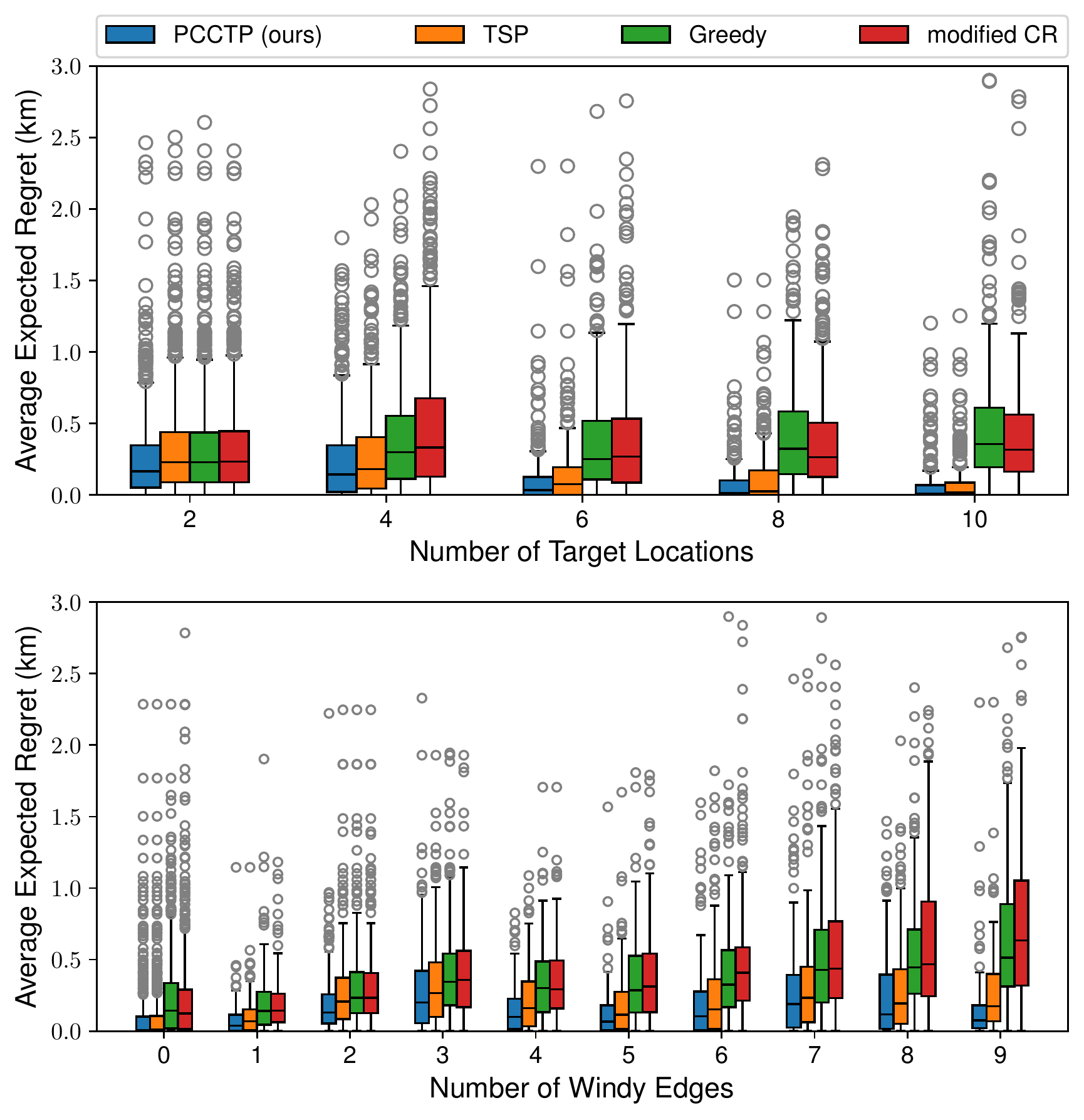}
    \caption{Average expected regret of PCCTP and baselines. A total of 2217 graphs are grouped according to the number of target nodes and windy edges. When there are no windy edges, all stochastic edges are pinch points.}
    \label{fig:sim_result}
\end{figure}

\vspace{-0.1cm}
\subsection{Baselines}
\vspace{-0.1cm}

The simplest baseline is an online greedy algorithm that always goes to the nearest unvisited target node assuming all ambiguous edges are traversable. 
For a graph with $k$ stochastic edges, we simulate all $2^k$ possible worlds, each with a different traversability permutation, and evaluate our greedy actor on each one.
The greedy actor recomputes a plan at every step and queries the simulator if it encounters a stochastic edge to disambiguate it.
Also, it checks the reachability of every target node upon discovering an untraversable edge and gives up on any unreachable targets.

A more sophisticated baseline is the optimistic TSP algorithm. 
Instead of always going to the nearest target node, it computes the optimal tour using dynamic programming to visit all remaining targets assuming all ambiguous edges are traversable.  
Similar to the greedy actor, TSP recomputes a tour at every step and may change its plan after it encounters an untraversable edge.
The expected cost is also computed via a weighted sum on all $2^k$ possible worlds. 
In contrast to PCCTP, both TSP and greedy require onboard computation to update their optimistic plans.

Lastly, we modify the CR algorithm, originally a method for CCTP \cite{cctp}, to solve PCCTP.
CR precomputes a cyclic sequence to visit all target nodes using the Christofides algorithm \cite{Christofides1976WorstCaseAO} and tries to visit all target nodes in multiple cycles while disambiguating stochastic edges.
If a target node turns out to be unreachable, we allow CR to skip this node in its traversal sequence. 


\vspace{-0.1cm}
\subsection{Results}
Fig. \ref{fig:sim_result} compares our algorithm against all baselines. 
To measure the performance across various graphs of different sizes, we use the average expected regret over all graphs. 
The expected regret of a policy $\pi$ for one graph $G$ is defined as
\begin{equation*}
\vspace{-0.2cm}
\mathbb{E}_w[\text{Regret} (\pi)] = \sum_{w} [ p(w)(\phi(\pi, w) - \phi(\pi^p, w)) ], \
\end{equation*}
where $\pi^p$ is a privileged planner with knowledge of the states of all stochastic edges, $\phi$ is the cost functional, and  $w$ is a possible world of the graph.
PCCTP precomputes the optimal policy in about 50 seconds on average in our evaluation, and there is no additional cost online.
Compared to the strongest baseline (TSP), our algorithm saves the robot about 1\%(50m) of travel distance on average and 15\%(1.8km) in the extreme case.
Although the improvement is marginal on average, our planner can still be beneficial in edge cases (e.g., high blocking probability, long stochastic edges). 
The performance of PCCTP may be further enhanced if the estimated blocking probabilities of the stochastic edges are refined based on historical data.
\begin{figure*}[t]
    \centering
    \includegraphics[width=0.94\linewidth]{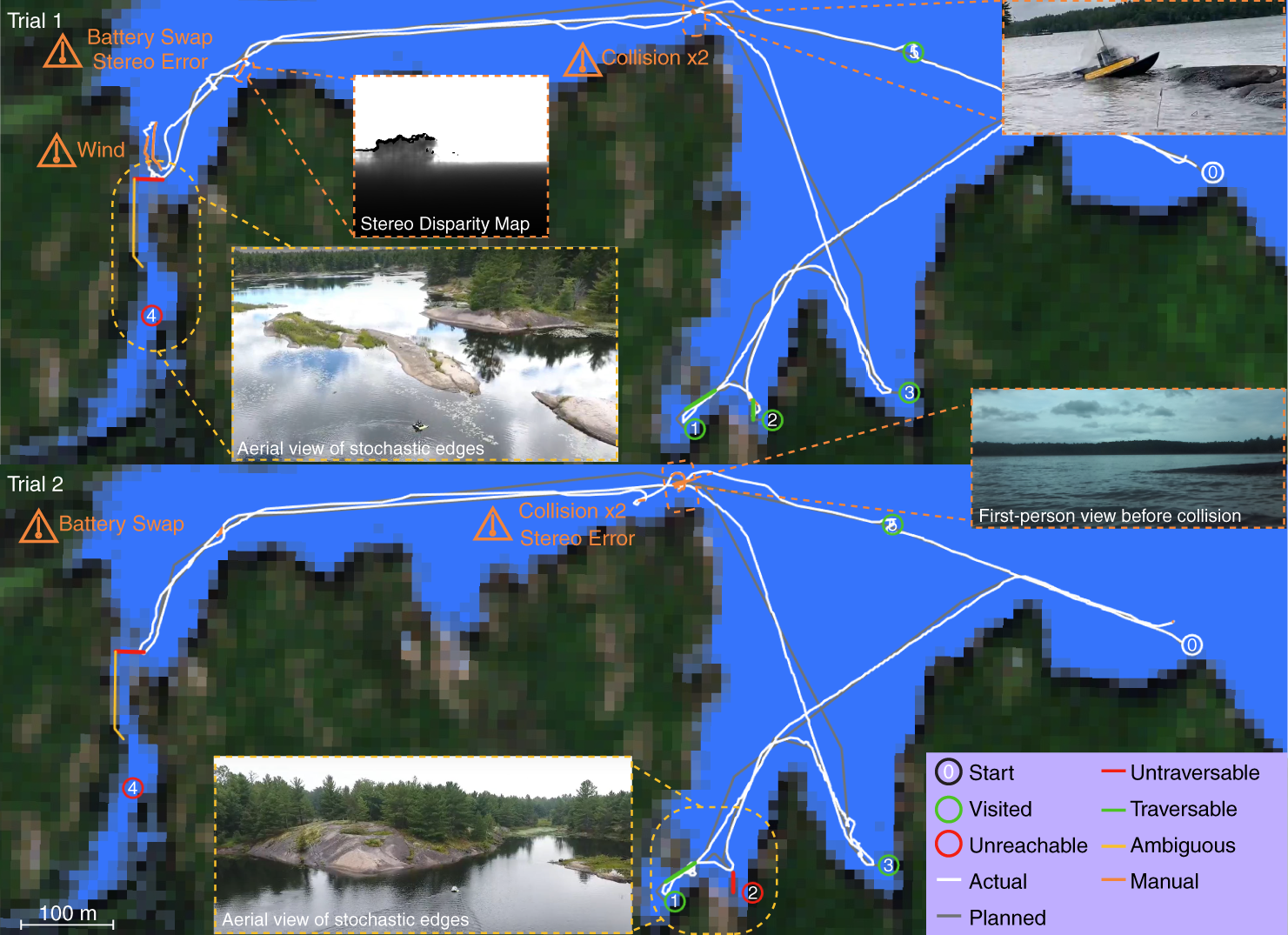}
    \caption{The GPS path of our robot on satellite water masks for two runs of our field test. 
    The total length traveled autonomously is about 3.7 km. 
    The pictures with a dashed \textcolor{Goldenrod}{yellow} border are drone aerial views of the stochastic edges.
    We have labeled all instances of manual interventions, including battery changes, failures to detect a rock, and a stereo matching failure in \textcolor{Orange}{orange}.
    Note that the labels untraversable, traversable, and ambiguous are the state of stochastic edges after disambiguation. } 
    \label{fig:path}
    \vspace{-0.5cm}
\end{figure*}

\section{Real Robot Experiment}
\vspace{-0.1cm}

\subsection{Robot System}

Our ASV platform, shown in Fig. \ref{fig:robot}, is a customized \textit{Clearpath Heron} ASV with a GPS, IMU, \textit{ZED} stereo camera,  \textit{Ping360} scanning sonar, and an \textit{Nvidia Jetson} for processing sensor inputs.
A separate \textit{Micro-Start XP-1} battery powers the \textit{Jetson}, stereo camera, and sonar for approximately one hour, 
while two NiMH cells power the motors and remaining electronics for about two hours. 
We did not integrate water samplers for this mission but we can add them in the future.  
Our system also includes a custom web interface that displays the status of the robot and the planner. 
We have a remote controller used for manual mode if needed for safety.

\vspace{-0.1cm}
\subsection{Autonomy Stack for Policy Execution}
Our philosophy is to design a robust
autonomy framework where the success of policy execution is less dependent on lower-level components such as perception and local planning.
At a high level, the planner precomputes the navigation policies from satellite images given user-selected sampling locations.
A local planner then follows the global path during a mission.
When the robot is disambiguating a stochastic edge, the policy executor will decide the traversability of the edge based on a timer.
If the robot finishes travelling the stochastic edge within the time limit, the edge is marked as traversable. Otherwise, the edge is untraversable.
The executor then branches into different policy cases depending on the disambiguation result.
Using a timer allows us to account for issues we cannot directly sense (e.g., heavy headwind), and does not require a perfect terrain assessment to identify a stochastic edge successfully. 

The local planner uses the Dynamic Window Algorithm (DWA) \cite{dwa} to plan a desired velocity that tracks the global plan and avoids obstacles. 
The robot's odometry and global pose are estimated with an extended Kalman filter that combines GPS and IMU data.
The local planner maintains a 2D rolling-window cost map that tracks obstacles in the local, 30 m by 30 m surroundings of our robot. 
We use a point-cloud-based method \cite{obsusv} to detect and track obstacles visible in the field of view of the stereo camera.
Obstacles are classified as clusters of points above the water plane and tracked using a temporal consistency filter.
The tracked obstacles are added or discarded only after consecutive detections for two seconds and used to update the cost map.
The sonar is only used for data collection and not for navigation in this work.

\vspace{-0.1cm}
\subsection{Testing Site}
We tested our algorithm at Nine Mile Lake in McDougall, Ontario, Canada, with five sampling locations of our choice.
An overview of the mission is shown in Fig. \ref{fig:path}.
To reach the target at the bottom-left of the satellite image, the ASV has to disambiguate three consecutive stochastic edges with several large rocks.
There are two more stochastic edges, each leading to a target node at the lower bay of the lake.
The aerial views of both sections are shown in Fig. \ref{fig:path}.
The large inconsistency between the satellite image and the local aerial view demonstrates the need for a robust algorithm. 
Our planner does exactly that by treating these narrow water passages as stochastic edges and plans accordingly. 

\vspace{-0.1cm}
\subsection{Results}
We conducted two trials: first on a rainy day and then on a sunny day. 
The high-level objective is to navigate the ASV to all reachable target nodes and return home. 
The robot successfully achieved that in both trials.
Manual interventions and their causes are labelled along the path in Fig. 
\ref{fig:path}.
Our robot did not find a traversable path to target node 4 because the global plan derived from satellite images was blocked by large rocks (see left of Fig. \ref{fig:path}) and the local planner failed to identify an alternative. 
We observed several other issues during the field tests: power, terrain assessment, and environmental disturbances. 
The \textit{XP-1} battery must be swapped after about an hour of use. 
Our stereo matching algorithm struggled against sun glare and very calm water. 
A shallow rock was misclassified as part of the water plane (see. right of Fig. \ref{fig:path}) and consistently resulted in collisions.
Lily pads damaged one of our underwater plastic propeller blades and very strong winds overpowered the motors at one point. 
Despite these partial system failures and disturbances, the ASV can still execute the full policy, visit other reachable target nodes, and return home.
This can be credited to the robustness of our framework, particularly because our policy execution only depends on the robot's global pose and an independent timer. 

\vspace{-0.2cm}
\section{Conclusion}
In this paper, we present a route-planning algorithm for ASV navigation using historical satellite images.
We propose to model the environmental disturbances as well as inaccuracies of the satellite images by building a stochastic roadmap of the target waterway. 
The route-planning problem can then be solved offline optimally by a heuristic-based tree-search algorithm. 
The correctness and utility of our planner are verified in a simulation of over a thousand lakes against two online planning baselines. 
We tested our algorithm in the field and showed how our planner can be part of a robust ASV navigation system. 

\section*{ACKNOWLEDGMENT}
We thank the Natural Sciences and Engineering Research Council of Canada (NSERC) for partially supporting this work.






\bibliographystyle{IEEEtran}
\bibliography{ref}

\end{document}